\documentclass[twocolumn]{article}

\usepackage[
  a4paper,
  left=20mm,
  right=20mm,
  top=20mm,
  bottom=20mm,
]{geometry}

\usepackage{times}
\usepackage{helvet}
\usepackage{courier}
\usepackage[hyphens]{url}
\usepackage{graphicx}
\usepackage{comment}
\usepackage{float}
\usepackage{authblk}

\usepackage[square,numbers]{natbib}

\usepackage{algorithm}
\usepackage{algorithmic}

\usepackage[
breaklinks=true,
colorlinks=true,
linkcolor=blue,citecolor=blue,urlcolor=blue,filecolor=blue,
pdffitwindow,
bookmarks=true,
bookmarksopen=true,
bookmarksnumbered=true,
pdftitle={Sequence learning, prediction, and replay in networks of spiking neurons},
pdfauthor={Younes Bouhadjar, Dirk J. Wouters, Markus Diesmann, Tom Tetzlaff},
pdfsubject={},
pdfkeywords={}
]
{hyperref}

\usepackage{xspace}

\usepackage{amsmath}
\usepackage{amssymb}
\usepackage{mathtools}

\usepackage{booktabs}
\usepackage{array}
\usepackage{cleveref}
\crefname{supp}{supplement}{Supplements}
\crefname{app}{appendix}{Appendices}
\crefformat{equation}{Eq~(#2#1#3)}
\crefformat{figure}{Fig.\,#2#1#3}
\crefformat{section}{Sec.\,#2#1#3}
\crefformat{appendix}{App.\,#2#1#3}
\crefformat{table}{Tab.\,#2#1#3}

\usepackage[commandnameprefix=ifneeded]{changes}

\definecolor{YB}{RGB}{0,0,200}
\definecolor{MF}{RGB}{0,200,0}
\definecolor{FS}{RGB}{200,0,0}
\definechangesauthor[color=YB]{YB}
\definechangesauthor[color=MF]{MF}
\definechangesauthor[color=FS]{FS}

\usepackage{pifont}
\usepackage{xcolor}

\newcommand{\cmark}{\textcolor{green!60!black}{\ding{51}}} 
\newcommand{\xmark}{\textcolor{red}{\ding{55}}}           

\newcommand{\panellabel}[1]{{\textbf{}#1}}  
\newcommand{\panel}[1]{{\textbf{}\panellabel{#1})}}  

\renewcommand{\exp}{\textnormal{exp}} 

\newcommand{\VocabSize}{$V$\xspace}
\newcommand{\AmbDepth}{$D$\xspace}
\newcommand{\NumAmb}{$A$\xspace}
\newcommand{\TransitionProb}{$p_\textnormal{T}$\xspace}
\newcommand{\GapProb}{$p_\textnormal{G}$\xspace}

\title{Dissecting Linear Recurrent Models: How Different Gating Strategies Drive Selectivity and Generalization}

\author[1]{Younes Bouhadjar\thanks{y.bouhadjar@fz-juelich.de}}
\author[1,2]{Maxime Fabre}
\author[1,3]{Felix Schmidt}
\author[1,4]{Emre Neftci}

\affil[1]{\footnotesize%
        Peter Grünberg Institute, Neuromorphic Software Ecosystems (PGI-15), Jülich Research Centre, Germany}
\affil[2]{Groningen Cognitive Systems and Materials Center (CogniGron), University of Groningen}
\affil[3]{\footnotesize%
        RWTH Aachen University, Aachen, Germany}
\affil[4]{\footnotesize%
        Department of Electrical Engineering and Information Technology, RWTH Aachen University, Aachen, Germany}

\date{\footnotesize\today}

\begin{document}

\maketitle


\begin{abstract}

Linear recurrent neural networks have emerged as efficient alternatives to the original Transformer's softmax attention mechanism, thanks to their highly parallelizable training and constant memory and computation requirements at inference.
Iterative refinements of these models have introduced an increasing number of architectural mechanisms, leading to increased complexity and computational costs.
Nevertheless, systematic direct comparisons among these models remain limited.
Existing benchmark tasks are either too simplistic to reveal substantial differences or excessively resource-intensive for experimentation.
In this work, we propose a refined taxonomy of linear recurrent models and introduce SelectivBench, a set of lightweight and customizable synthetic benchmark tasks for systematically evaluating sequence models. SelectivBench specifically evaluates selectivity in sequence models at small to medium scale, such as the capacity to focus on relevant inputs while ignoring context-based distractors. It employs rule-based grammars to generate sequences with adjustable complexity, incorporating irregular gaps that intentionally violate transition rules.
Evaluations of linear recurrent models on SelectivBench reveal performance patterns consistent with results from large-scale language tasks. Our analysis clarifies the roles of essential architectural features: gating and rapid forgetting mechanisms facilitate recall, in-state channel mixing is unnecessary for selectivity, but critical for generalization, and softmax attention remains dominant due to its memory capacity scaling with sequence length.
Our benchmark enables targeted, efficient exploration of linear recurrent models and provides a controlled setting for studying behaviors observed in large-scale evaluations.
Code is available at \url{https://github.com/symseqbench/selectivbench}

\end{abstract}



\section{Introduction}

Linear recurrent models have shown promising results as efficient alternatives to softmax attention due to their parallelizability and linear-time inference. What started from state space models \citep{Gu21_S4} or linearized attention \citep{Katharopoulos20_linear_transformers} developed into more complex models introducing data-dependent gating \citep{Gu23_mamba, Yang23_gla}, matmul state transition \citep{Yang24_delta_net}, multi-layer perceptron states \citep{behrouz2024titans, behrouz2025atlas}, or multi-steps updates \citep{schone2025implicit, von2025mesanet, Siems25_delta_product}, always trying to push their performance further. But despite their growing use, a clear unified taxonomy and direct comparisons across different architectural variants remain limited, in part due to the lack of diagnostic, lightweight benchmarks. Prototyping on NLP tasks is often impractical: it demands extensive resources, long training cycles, and fails to isolate specific model capabilities. Synthetic benchmarks, such as RULER \citep{Hsieh24_ruler} and LongBench \citep{Bai23_longbench}, offer more targeted evaluations but still require full pretraining and remain computationally expensive.
\par
On the other hand, recent works have proposed smaller-scale synthetic benchmarks to reduce computational costs. For instance, the MQAR benchmark \citep{Arora23_zoology} has been a central benchmark for the development of linear recurrent models \citep{Yang24_delta_net, Yang23_gla} thanks to its evaluation of memorization or span recall capacities. 
Nevertheless, it has been omitted from more recent model developments \citep{Yang24_gated_delta_net, Siems25_delta_product} due to its limited benchmarking relevance, stemming from the use of overly simplistic rhythmic sequences composed of key-value pairs.
Another common approach involves sequences generated based on principles from formal language theory \citep{Deletang22_neural, Akyurek24_ICLL}. While some of these works are quite theoretical and do not consider the real challenges of large scale models, the RegBench benchmark \citep{Akyurek24_ICLL} has come up as a realistic and popular tool for prototyping LLM models.
RegBench focuses on evaluating in-context learning by training models on regular languages generated from random finite automata. However, it does not provide a quantitative measure of grammar complexity, nor does it introduce gaps between sequence items, both of which are critical for assessing selectivity in sequence learning models.
\par
In this work, we propose SelectivBench, a set of benchmark tasks that extends SymSeqBench \citep{SymSeqBench} to test small-scale language models on a range of tasks of growing complexity, evaluating NLP-like advanced features required for attention models such as selectivity.
To further challenge the selectivity aspect of these models, we introduce gap elements into the generated sequences, either as noise or as vocabulary items that violate the transition rules. The latter requires the models to rely on contextual information to distinguish valid patterns. This setup offers a precise means of assessing models' selective processing capabilities. The final task of SelectivBench evaluates the capacity of models to generalize beyond their training sequence length.
We first establish a unified categorization of linear recurrent models, focusing on their selectivity features. Then, by systematically varying both the complexity of the grammar and the level of selectivity, we demonstrate the role of various sequence processing mechanisms, such as gating and channel mixing, in model performance across varying task demands.
Our evaluations show that gating and fast decay mechanisms aid recall, while in-state channel mixing is essential for generalization. Despite the efficiency of these models, softmax attention remains superior in tasks requiring extended memory.
The results support further development of models that incorporate or extend such mechanisms, and encourage exploration of new architectural strategies to better handle distractors in structured sequences.

\section{Related Work}

Language models are typically pretrained on a large corpus of text (e.g., FineWeb \cite{Penedo24_FineWeb}) and then evaluated zero-shot on tasks such as reasoning \citep{Weston14_MemoryNetworks}, retrieval \citep{Arora24_just_read_twice}, and long-context understanding \citep{Bai23_longbench}. While effective, these benchmarks are costly and offer limited insight into specific architectural features. Synthetic tasks such as RULER \cite{Hsieh24_ruler} and NeedleInAHaystack \cite{Kamradt23_nihs} provide more targeted diagnostics but still depend on large models and high compute, limiting their use in early-stage analysis.
\par
Lightweight synthetic tasks offer a more efficient alternative for targeted evaluation. Early benchmarks such as those proposed in \cite{Graves14_ntm, Weston14_MemoryNetworks} 
tested limited capabilities like span recall and copying.
The MQAR task \cite{Arora23_zoology}, which relies on simple sequences of keys and value pairs, has been widely used to probe associative recall in small language models. Despite their efficiency, these tasks are easy to solve and provide limited insight into the performance of models at scale, due to their artificial sequence structure and their lack of evaluation of the selective processing abilities of these models.
\par
A growing body of work has explored the use of formal languages to construct structured input sequences that better capture the hierarchical and rule-based characteristics of natural language \citep{White21_inductive_bias_ag, AllenZhu25_physicslanguagemodels}.
Among these efforts, RegBench \cite{Akyurek24_ICLL} has been widely adopted to evaluate linear recurrent models, particularly focusing on their in-context learning abilities.
It constructs multiple AGs with varying parameter combinations and trains models jointly across them. At test time, models are evaluated on data generated from a previously unseen grammar.
However, RegBench lacks explicit control over the complexity of the underlying grammars and does not incorporate gaps between sequence elements, which limits its effectiveness for evaluating selective processing in sequence models.
More recently, the Associative Tree Recall (ATR) benchmark \citep{Arora25_atr} extends associative recall to hierarchical settings by generating sequences from a probabilistic context-free grammar (PCFG), requiring retrieval over non-adjacent, tree-structured dependencies.
While ATR is more challenging and more structurally similar to natural language than, for instance, the MQAR task, it still offers only coarse control over grammatical complexity and does not explicitly study the influence of distractor symbols.
\par
Recently, SymSeqBench \citep{SymSeqBench} enabled more refined evaluations by exploring grammars with explicit and quantifiable complexity. Our proposed set of benchmark tasks, SelectivBench, extends on SymSeqBench by systematically increasing the required selective capacity of models through controlled disruptions. This design enables more granular analysis of model behavior and helps isolate the role of architectural components, such as gating and memory mechanisms, in generalization and selective processing.


\begin{figure}[!h]
  \centering
  \includegraphics[width=\linewidth]{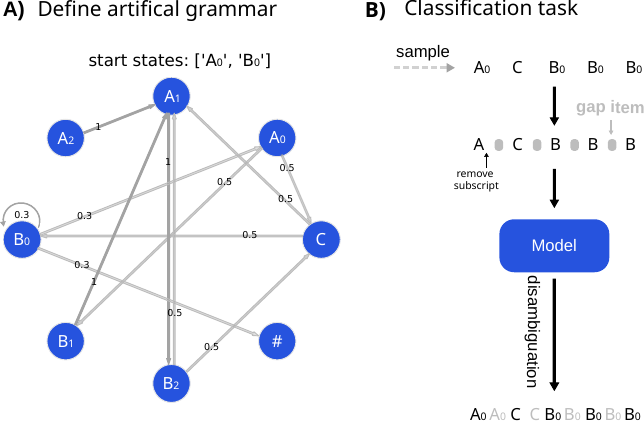}
  \caption{%
    \textbf{Illustration of the benchmark task.} \panel{A} The diagram defines the AG rules used to generate sequences.
    \panel{B} An illustration of a generated sample and the experimental setup (see \nameref{sec:artifical_grammer} section for more details.) 
  }
\label{fig:grammar} 
\end{figure}

\section{Methods}

\subsection{Models}
\label{sec:models}

The rise of large language models (LLMs) and long-sequence processing in natural language processing (NLP) has been driven largely by the success of the Transformer architecture \citep{Vaswani17_attention}. Transformers remain the de facto standard for a wide range of NLP tasks due to their ability to model complex, long-range dependencies through the self-attention mechanism.
However, the standard softmax-based attention comes with high computational and memory costs: during training, its complexity scales quadratically with the sequence length due to the full pairwise attention score computation, and at inference time, token-by-token generation requires a key-value cache and computations that grow linearly with sequence length. These implementation costs of softmax attention pose serious challenges when scaling to longer sequences or deploying models in resource-constrained settings.

\paragraph{Linear Recurrent Memories.}
More recently, a new family of models based on linear recurrent dynamics has emerged as a promising, efficient alternative to transformers. Drawing on ideas from state space modeling (S4 \cite{gu2022parameterization}, HiPPO \cite{Gu20_hippo}, S5 \cite{Smith23_S5}, LRU \cite{Orvieto23_lru}) or from linearized softmax attention \citep{Katharopoulos20_linear_transformers}, several approaches have converged on closely related neural memory dynamics that perform well on long sequence processing tasks. This class of models has been described as linear attention, state space models, or linear recurrent neural networks. Despite differences in terminology, they share a common design principle: a finite memory or state $S$ that is updated through a linear recurrent mechanism with element-wise transitions:
\begin{equation}
  S_{t} = A \odot S_{t-1} + B_t \otimes I_t,
\end{equation}
before being read out:
\begin{equation}
  y_t = S_{t} \times C_t,
\end{equation}
where $y$ is the output of the model.
We therefore adopt a more general unifying term: linear recurrent memories (LRM).
Here, the transition matrix $A$ acts element-wise through $\odot$. At each timestep, the input $I_t$, obtained by projecting the token $x_t$, is expanded from dimension $d$ to $d \times N$ by the outer product $B_t \otimes I_t$. The resulting state in $\mathbb{R}^{d \times N}$ is read out using $C_t$ through a matrix--vector product $(\times)$, equivalently an element-wise product followed by a sum over the $N$ dimension, yielding $y_t \in \mathbb{R}^d$.
\par
These models offer a significantly reduced computation cost and memory requirement due to their finite recurrent state, while still allowing for fast sequence length-parallelized training thanks to their linear nature. 

\begin{table*}[!h]
\centering
\begin{tabular}{l|cccc}
\hline
\textbf{Model} & \textbf{Selectivity} & \shortstack{\textbf{Complementary} \\ \textbf{Gating}} & \shortstack{\textbf{Channel} \\ \textbf{Mixing}} & \shortstack{$\#$ \textbf{Gate} \\ \textbf{Parameters}} \\
\hline
L.A. \citep{Katharopoulos20_linear_transformers} & \xmark & \xmark & \xmark & 0 \\
S4D \citep{gu2022parameterization}              & \xmark & \xmark & \xmark & $d \times d_{\text{state}}$ \\
DeltaNet \citep{Yang24_delta_net}      & $\boldsymbol{\sim}^*$ & \cmark & \cmark & $d \times N_{\text{heads}} + d^2$ \\
GLA \citep{Yang23_gla}            & \cmark & \xmark & \xmark & $\sim d \times 16$ \\
Mamba \citep{Gu23_mamba}           & \cmark & \cmark & \xmark & $2d \times d_{\text{state}} + 8 d^2/16$ \\
Mamba2 \citep{Dao24_mamba2}         & \cmark & \cmark & \xmark & $\sim 2d \times N_{\text{heads}}$ \\
Gated DeltaNet \citep{Yang24_gated_delta_net} & \cmark & \cmark & \cmark & $\sim 2d \times N_{\text{heads}} + d^2$ \\
DeltaProduct \citep{Siems25_delta_product} & \cmark & \cmark & \cmark & $n_h \times (d \times N_{\text{heads}} + d^2)$ \\
\hline
\end{tabular}
\\
{\footnotesize $^{*}$Weak selectivity: this model's state transition matrix $A_t$ doesn't allow for fast forgetting.}
\caption{Comparison of gate parameterization and gating features across linear recurrent memory (LRM) models.} 
\label{tab:mod_cats}
\end{table*}

\paragraph{Selective Linear Recurrent Memories.}

Early state-space models such as LMU \cite{Voelker19_LMU}, LRU \cite{Orvieto23_lru}, S4 \cite{Gu21_S4}, as well as the Linear Attention model \cite{Katharopoulos20_linear_transformers}, exhibit limited performance on complex tasks, in particular in natural language processing.
The performance gap compared to transformer models has been partially addressed by more effective selectivity mechanisms, such as those in Mamba \citep{Gu23_mamba}, enabling the state to retain relevant inputs while discarding the rest.
In LRM models, this feature can, for instance, be implemented by making the state transition matrix $A$ data-dependent.
\begin{equation}
A_{t} = f(a, g_t), \quad \textnormal{with} \, \, \, g_t = W_\text{G} x_t,
\end{equation}
where $f$ is a nonlinear function of the parameter $a$ and the gate projection $g_t$.
Common choices include $f(x,y) = x \cdot \sigma(y)$ or $f(x,y) = \exp\big(-x \cdot \zeta(y)\big)$, where $\sigma$ denotes the sigmoid function and $\zeta$ the softplus function. Additionally, such selective LRM models rely on a data-dependent input and readout gating $B_t$ and $C_t$ defined similarly to $A_t$.
\par
A whole range of selective models with element-wise linear recurrence have since been developed: GLA \citep{Yang23_gla}, Mamba2 \citep{Dao24_mamba2}, HGRN and HGRN2 \citep{Qin24_hgrn2}, RWKV6 and RWKV7 \citep{Peng23_rwkv}, mLSTM \citep{Beck24_xlstm}, Griffin \citep{De24_hawk_griffin}, Rodimus \citep{He25_rodimus}. Although these models can present multiple variations, like the choice of the $f$ function, we highlight here a key performance-impacting design choice: while models like GLA \citep{Yang23_gla} rely on two completely independent $A_t$ and $B_t$ gates, others like Mamba \citep{Gu23_mamba} display a complementary mechanism between these two gates such that with increasing gate $g_t$, $A_t = \exp(-a \cdot \zeta(g_t))$ decreases while $B_t = \zeta(g_t) \cdot W_\text{B} x_t$ increases, and vice-versa.
This coupling leads to a coordinated behavior: new information is written when forgetting occurs, and stored information is preserved when input is erased. We refer to this as complementary gating and examine its effects in detail in the \nameref{sec:experiments} section.

\paragraph{Channel-mixing in Linear Recurrent Memories.}

Another attempt to improve the performance of LRMs was proposed by DeltaNet \citep{Yang24_delta_net}. While models introduced above rely on an element-wise product between $A_t$ and $S_t$ for the state update, only mixing information between successive tokens in time, DeltaNet, inspired by the work of \citet{Schlag21_fwm, Schlag21_linear_transformers}, instead employs a matrix-matrix multiplication to model more complex memory interactions, introducing an extra mix between different channels of the memory. To afford this channel-mixing without paying the prohibitive cost of a full matrix-matrix multiplication, such LRMs employ a diagonal-plus-low-rank transition matrix
$
A_t=\mathbf{I} - \sigma(h_t)k_tk_t^\top
$.
This allows to maintain an efficient parallelization kernel, around $10\%$ slower than the element-wise-based GLA kernels (see Figure 6 of \citep{Yang24_delta_net}). This feature was also implemented more recently in the TTT model \citep{sun2024learning}.
\par
DeltaNet's input gating $B_t = \sigma(g_t) k_t$, coupled to the transition $A_t$, ensures complementary gating. Nevertheless, DeltaNet's definition of $A_t$ prevents it from approaching 0, as $k_t$ is normalized and its norm is strictly below 1. This directly hinders the model from erasing memory rapidly. We refer to this as weak selectivity and examine its consequences in the \nameref{sec:experiments} section.
To address this limitation, the Gated DeltaNet \citep{Yang24_gated_delta_net} inspired by Mamba2 introduces an additional projection gate $\alpha_t$, which results in the following transition matrix
$
A_t = a \cdot \zeta(\alpha_t)\, \left( \mathbf{I} - \sigma(g_t) k_t k_t^\top \right),
$
enabling a fast clearing of the state.

\paragraph{Rank-n Linear Recurrent Memories.}

More recently, several works have extended the long-context learning capabilities of LRMs by deliberately increasing the complexity and thus the computational cost of the state update. This includes works such as TTT \citep{sun2024learning}, Titans \citep{behrouz2024titans}, and Atlas \citep{behrouz2025atlas}, which either replace the recurrent state with a multi-layer perceptron or augment LRMs with sliding-window attention to support context recall.
Other approaches to increase state-update complexity instead target the rank of the state transition matrix. MesaNet \citep{von2025mesanet} follows this strategy by introducing an additional recurrent state $H_t$ involved in the final readout, while implicit SSMs \citep{schone2025implicit} rely on fixed-point self-iterations within the state update. DeltaProduct \citep{Siems25_delta_product} pushes this idea further by extending the DeltaNet transition with a product of $n_h$ Householder matrices derived from independent gates,
\begin{equation}
A_t = a \cdot \zeta(\alpha_t) \prod_{i=1}^{n_h} \left(\mathbf{I} - \sigma(g_{t,i})\, k_{t,i} k_{t,i}^\top\right).
\end{equation}
This construction increases the effective rank of the transition, enabling channel rotations in addition to DeltaNet's reflective operations, at the cost of a throughput reduction of roughly $50\%$ for $n_h = 4$.
\par
We summarize the established categories and features of LRMs in \cref{tab:mod_cats} for the most relevant models in each category.
\cref{tab:models} further reports the architectures and sizes of the LRM models studied in this work and evaluated in \nameref{sec:experiments}, except for Linear Attention and S4D, which are included only for reference.
\cref{tab:gate_params_summary} further details the number of parameters in the state-transition gate $A_t$ for the largest model configurations ($d=1296$, $N_{\text{heads}}=16$, $d_{\text{head}}=81$, $d_{\text{state}}=64$ for Mamba and $128$ for Mamba2).

\begin{table*}[!t]
\centering
\begin{tabular}{lcc}
\hline
\textbf{Model Name} & \textbf{\# Gate Params} & \textbf{\# Gate Params in Largest Used Config} \\
\hline
DeltaNet \citep{Yang24_delta_net} & $d \times N_{\text{heads}} + d^2$ & 1.7M \\
GLA \citep{Yang23_gla} & $\sim d \times 16$ & 21k \\
Mamba \citep{Gu23_mamba} & $2d \times d_{\text{state}} + \frac{8 d^2}{16}$ & 1M \\
Mamba2 \citep{Dao24_mamba2} & $\sim 2d \times N_{\text{heads}}$ & 41k \\
Gated DeltaNet (Yang et al. 2025) & $\sim 2d \times N_{\text{heads}} + d^2$ & 1.7M \\
DeltaProduct \citep{Siems25_delta_product} & $n_h \times (d \times N_{\text{heads}} + d^2)$ & 6.7M \\
\hline
\end{tabular}
\caption{Number of gate parameters for transition matrix $A_t$ and effective values for the largest configurations of all studied LRM models ($d$=1296).}
\label{tab:gate_params_summary}
\end{table*}

\begin{table*}[t]
\centering
\renewcommand{\arraystretch}{1.4}
{\small
\begin{tabular}{
    >{\centering\arraybackslash}m{3cm}|
    >{\centering\arraybackslash}m{3.5cm}
    >{\centering\arraybackslash}m{2.5cm}
    >{\centering\arraybackslash}m{2.5cm}
    >{\centering\arraybackslash}m{3cm}
}
\toprule
\textbf{Model} & \(A_t\) & \(B_t\) & \shortstack{State Size \\ ($d=1296$ config.)} & \multicolumn{1}{c}{Learnable Parameters} \\
\midrule
L.A. \citep{Katharopoulos20_linear_transformers} &
\( \mathbf{I}\)
&
\( k_t \) &
\shortstack{\( d^2/N_{\text{heads}} \) \\ \( (105k) \)} &
0
\\
S4D \citep{gu2022parameterization} &
\( \exp(-A) \) &
\( \mathbf{I}\) &
\shortstack{\( d_{\text{state}} \times d \) \\ \( (83k) \)} & $A \in \mathbb{R}^{d_{\text{state}} \times d}$
\\ 
DeltaNet \citep{Yang24_delta_net} &
\( \mathbf{I} - \sigma(g_t)k_tk_t^\top \) &
\( \sigma(g_t)k_t \) &
\shortstack{\( d^2/N_{\text{heads}} \) \\ \( (105k) \)} &
$W_\text{g} \in \mathbb{R}^{d \times N_{\text{heads}}} $
\\
GLA \citep{Yang23_gla} &
\( \sigma(g_t)^{1/\tau} \) &
\( k_t\) &
\shortstack{\( d^2/2N_{\text{heads}} \) \\ \( (52k) \)}&
\( W_\text{g1} \in \mathbb{R}^{d \times 16} \), 
\( W_\text{g2} \in \mathbb{R}^{16 \times d/2N_{\text{heads}}} \)
\\
Mamba \citep{Gu23_mamba} &
\( \exp(-A  \zeta(g_t))  \) &
\( \zeta(g_t)  W_B v_t  \) &
\shortstack{\( d_{\text{state}} \times 2d \) \\ \( (166k) \)}
 & 
\( A \in \mathbb{R}^{2d \times d_{\text{state}}} \), 
\( W_\text{g1} \in \mathbb{R}^{2d \times \frac{2d}{16}} \), 
\( W_\text{g2} \in \mathbb{R}^{\frac{2d}{16} \times 2d} \),
$ W_B \in \mathbb{R}^{2d \times d_{\text{state}}} $
\\
Mamba2 \citep{Dao24_mamba2} &
\( \exp(-a \zeta(g_t)) \)
&
\( \zeta(g_t) k_t  \)
&
\shortstack{\( d_{\text{state}} \times 2d \) \\ \( (332k) \)} &
\( a \in \mathbb{R}^{N_{\text{heads}}} \),
\( W_\text{g} \in \mathbb{R}^{2d \times N_{\text{heads}}} \)
\\
Gated DeltaNet \citep{Yang24_gated_delta_net} &
\( a \zeta(\alpha_t)(\mathbf{I} - \sigma(g_t)k_tk_t^\top) \) &
\( \sigma(g_t)k_t \) &
\shortstack{\( 2d_{\text{head}}^2 \times N_{\text{heads}} \) \\ \( (210k) \)}&
$ a \in \mathbb{R}^{N_{\text{heads}}} $, $W_{\alpha} \in \mathbb{R}^{d \times N_{\text{heads}}}$, $W_\text{g} \in \mathbb{R}^{d \times N_{\text{heads}}}$
\\
Gated DeltaProduct \citep{Siems25_delta_product} &
\( a \zeta(\alpha_t) \) \newline \( \prod_i^{n_h}{(\mathbf{I} - \sigma(g_{t,i})k_{t,i}k_{t,i}^\top)} \) &
$ \sum_k^{n_h}{\Big( \prod_{i=k+1}^{n_h}} $
 $ (\mathbf{I} - \sigma(g_{t,i})k_{t,i}k_{t,i}^\top) \Big) $
 $ \sigma(g_{t,k})k_{t,k} $
&
\shortstack{\( 2d_{\text{head}}^2 \times N_{\text{heads}} \) \\ \( (210k) \)}&
$ a \in \mathbb{R}^{N_{\text{heads}}} $, $W_{\alpha} \in \mathbb{R}^{d \times N_{\text{heads}}}$, $\{W_{h,(i)}\} \in \mathbb{R}^{d \times N_{\text{heads}} \times n_h}$
\\
\bottomrule
\end{tabular}

\caption{State-space formulations of attention and SSM-based sequence models, comparing the time-varying transition $A_t$, input $B_t$, resulting state dimensionality for $d=1296$, and learnable parameter counts across Linear Attention, S4D, DeltaNet, GLA, Mamba, Mamba2, and gated Delta variants.
For simplicity, all multiplications are written by juxtaposition. 
Depending on context, this denotes a dot product, matrix multiplication, or element-wise product. 
The learnable parameters for the standard $k$, $q$, $v$ projections are omitted here for compactness, as all models use them.
The only exception is S4D, which has its input gating $B_t$ independent of $k_t$ or $v_t$.
We also report the effective state size for the largest configuration considered. 
When two gate matrices $W_{g1}$ and $W_{g2}$ are shown, the gate is obtained via two successive projections, $g_t = W_{g2} W_{g1} x_t$.
}
\label{tab:models}}
\end{table*}

\subsection{Artificial Grammars}
\label{sec:artifical_grammer}

The SymSeqBench (SSB) framework \cite{SymSeqBench} provides tools for generating sequences from predefined artificial grammars (AGs). To clarify our proposed extension, we formalize the AGs used in SSB via a Markov chain (MC) representation.
A set of observable states \( \mathcal{S} \) and latent states \( \mathcal{Z} \) are defined, such that \( \mathcal{S} \) partitions \( \mathcal{Z} \) into disjoint groups. A sequence is generated via a Markov chain over \( \mathcal{Z} \), but the model only observes the representatives in \( \mathcal{S} \). Thus, an observed state \( s \in \mathcal{S} \) may correspond to any latent state \( z \in s \).
We denote by \( A(s) = |s| \) the number of latent states (number of ambiguities) associated with \( s \); in our experiments, this number is fixed for all \( s \in \mathcal{S} \) and is denoted by \NumAmb.
Depending on context, \( s \) may be called a partition element, observable state, or ambiguous region (\cref{fig:grammar}A).
SSB introduces ambiguity depth (\AmbDepth) as the size of the largest partition (i.e., the maximum number of latent states mapping to a single observable state), while the vocabulary size (\VocabSize) is \( |\mathcal{Z}| \).
To handle sequence termination, SSB includes a terminal state \( \# \), extending \( \mathcal{Z} \) and \( \mathcal{S} \) to \( \hat{\mathcal{Z}} = \mathcal{Z} \cup \{\#\} \) and \( \hat{\mathcal{S}} = \mathcal{S} \cup \{\#\} \). The terminal state is unambiguous and fully observable.
Sampling from the MC results in generated sequences of the form $z_{1:T} = (z_1, z_2, \dots, z_T)$, where $z_t \in \hat{\mathcal{Z}}$ with observable equivalents $s_{1:T} = (s_1, s_2, \dots, \hat{s}_T)$ with $s_t \in \hat{\mathcal{S}}$.
Each element in the resulting sequence is one-hot encoded over the set \( \hat{\mathcal{S}} \).
Finally, SSB provides several measures of sequence complexity, including the topological entropy (TE) introduced in \cite{Bollt20_artificial_grammars}, which we use in this work. TE quantifies the exponential growth rate of the number of distinct strings, thereby capturing increasing complexity as the generative grammar becomes richer \citep{Robinson98, Warren15_topological}.
In practice, TE is computed as the logarithm of the largest real eigenvalue of the grammar's Boolean (topological) state-transition matrix, obtained by binarizing its probabilistic transition matrix.
\par
In this work, we extend the SSB framework to evaluate the sequence processing capabilities of recent linear recurrent models as outlined in the following.

\subsection{SelectivBench}
\label{sec:data_generation}

We now present SelectivBench, a novel set of benchmark tasks, extending SymSeqBench's artificial grammar and sequence generation, to evaluate different key properties of small-scale language models, especially their selectivity capacities. The benchmark consists of four tasks, each scaling up in complexity to evaluate more advanced features of the models. As it is defined based on the SymSeqBench configurable artificial grammar, SelectivBench is configurable in complexity, but here we adopt a specific configuration that better differentiates the capacities of the evaluated models (see \nameref{sec:experiments} for details).

\paragraph{Task 1: Memorization for Disambiguation.}

Causal attention units, including softmax attention and linear recurrent memories (LRM), are designed to store information over a sequence and to retrieve the relevant content in the appropriate context.
To this end, \textbf{Task 1} evaluates the ability of models to learn the AG rules by testing whether they can disambiguate the observable states $s$ at a given position based on context. This capacity for memorization and disambiguation is fundamental to natural language processing (NLP) and therefore central to the models we study.
\par
For this task and all others, the following data pipeline is employed:
\begin{enumerate}
    \item Select a complexity level and generate a corresponding set of AG rules.
    \item Generate a set of train and test sequences $z_{1:T} = (z_1, z_2, \dots, z_T)$ of max length $T$ from this grammar.
    \item Prepare ambiguous sequences $s_{1:T} = (s_1, s_2, \dots, \hat{s}_T)$ by removing their state indices.
    \item Train and test a model $f$ to process ambiguous sequences, $f(s_{1:T}) = y_{1:T}, \quad \text{with } y_{1:T} = z_{1:T}$.

\end{enumerate}

Specifically, for each new ambiguous element in the sequence, the model has to classify the corresponding latent state based on the context. The performance is evaluated as the accuracy of the classified state during presentation of its ambiguous counterpart, averaged over the sequence length $\sum_{t=1}^T{\delta(y_t-z_t)}/T$. We also enforce constraints on AG rules to ensure that all sampled trajectories can be disambiguated by a model, allowing the reference accuracy to reach up to 100\%.

\paragraph{Task 2: Noise Rejection Selectivity.}
While memorization is important for sequence modeling, exact storage of all tokens is inefficient, particularly for fixed-state models such as LRMs. As highlighted in \citep{Gu23_mamba}, the central challenge for these models is to compress the context into a limited state by retaining relevant information and discarding or forgetting irrelevant content. This ability is referred to as selectivity.
\par
\textbf{Task 2} in SelectivBench inserts noise tokens, or gaps, between the generated sequence elements to artificially increase the required disambiguation context window and promote models to be selective. This is done while maintaining the natural and complex sequence rhythm obtained with the artificial grammar.
In practice, we extend \textbf{Task 1} sequences with noise gaps of varying durations:
$$s_{1:T} = (s_1, \mathbf{e}_1^{(1:n_1)}, s_2, \mathbf{e}_2^{(1:n_2)}, \dots, s_T, \mathbf{e}_T^{(1:n_T)}),$$
where $\mathbf{e}_t^{(1:n)} = (e_{t,1}, e_{t,2}, \dots, e_{t,n})$.
The gap durations $n$ are sampled from a uniform distribution during training between $n_\text{min}$ and $n_\text{max}$ and fixed during test.
While sequence elements $s_t$ are encoded as one-hot vectors, the noise tokens $e_{t,i}$ are represented by dense vectors in $\mathbb{R}^{|\hat{\mathcal{S}}|}$ whose components are drawn independently from a uniform distribution on $[0, \gamma]$.
The model's task is to disambiguate the sequence by ignoring the noise gaps and to output:
$
y_{1:T} = (z_1, \mathbf{z}_1^{(1:n_1)}, z_2, \mathbf{z}_2^{(1:n_2)}, \dots, z_T, \mathbf{z}_T^{(1:n_T)}),
$
where $\mathbf{z}_t^{(1:n)} = (z_{t,1}, \dots, z_{t,n})$ denotes $n$ repetitions of $z_t$.

\paragraph{Task 3: Context-aware Selectivity.}
The noise gaps in \textbf{Task 2} require selectivity to avoid memory overload, but they are relatively easy to ignore. Models can simply learn to classify noise tokens independently of context and filter them out.
To provide a more challenging selectivity evaluation, \textbf{Task 3} introduces gaps composed of non-grammatical elements, that is, elements that are not reachable from the current position in the grammar transition graph. This tests whether models can express selectivity relative to the specific context and therefore manage memory compression more effectively.
In practice, we introduce gap elements $e_t$ drawn from the set of states that are non-reachable from the current observable state $z_t$. Formally, let $\bar{\mathcal{S}}(z_t)$ denote the set of all $\bar{s} \in \mathcal{S}$ such that for every $z^{\prime} \in \bar{s}$ we have $\tau(z_t, z^{\prime}) = 0$, where $\tau(z_t, z')$ denotes the transition probability from $z_t$ to $z'$. We then sample $e_t$ uniformly from $\bar{\mathcal{S}}(z_t)$.
These non-grammatical gap elements are encoded in the same one-hot representation as regular sequence symbols.
Given the difficulty of this task, we restrict each gap to at most one token and instead control the presence of gaps using a per-position gap probability \GapProb.

\paragraph{Task 4: Length Generalization.}

\textbf{Task 4} assesses the ability of models to generalize their selectivity and memory handling to sequences and contexts that are longer than those seen during training. Models are trained with noisy gaps as in \textbf{Task 2} and evaluated on progressively larger gap lengths that exceed the training length.
\par
Length generalization is crucial for enabling large language models to operate effectively on very long contexts while keeping training costs manageable with shorter sequences. Transformers are known to struggle with such extrapolation, particularly under standard positional encodings \citep{zhou2024transformers}, and typically require at least partial long-context training through approaches such as curriculum learning \citep{awasthi2023improving}. Here, we benchmark softmax attention and LRM models without any specialized training protocol, in order to evaluate the capabilities of the core architectures themselves.


\begin{figure*}[!h]
    \centering
    \includegraphics[width=6in]{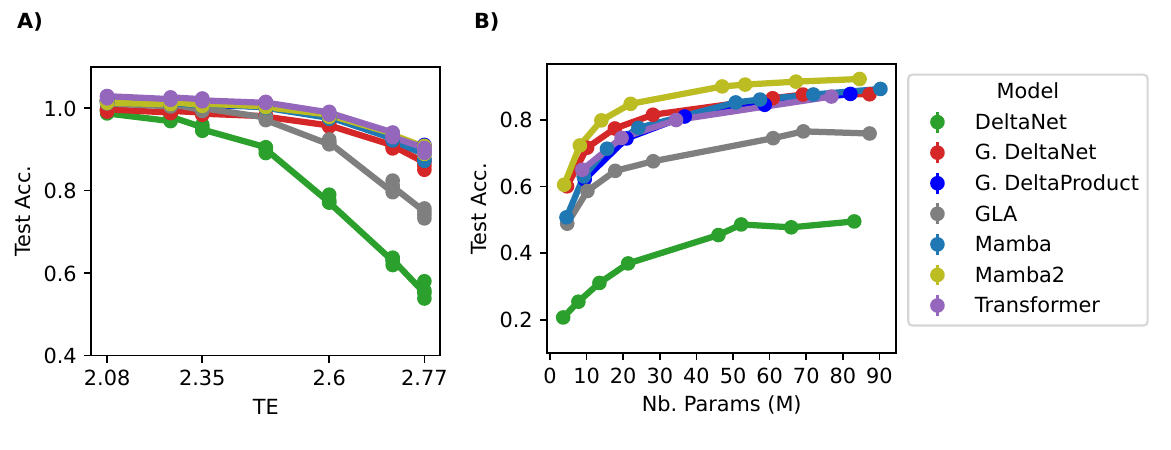}
    \caption{Test accuracy as a function of the sequence complexity level (TE) (A) and the model size (number of parameters in millions) (B).
    Results are averaged over 4 runs and reported as mean $\pm$ standard deviation.}
    \label{fig:acc_complexity_model_size}
\end{figure*}

\begin{figure*}[!h]
    \centering
    \includegraphics[width=6in]{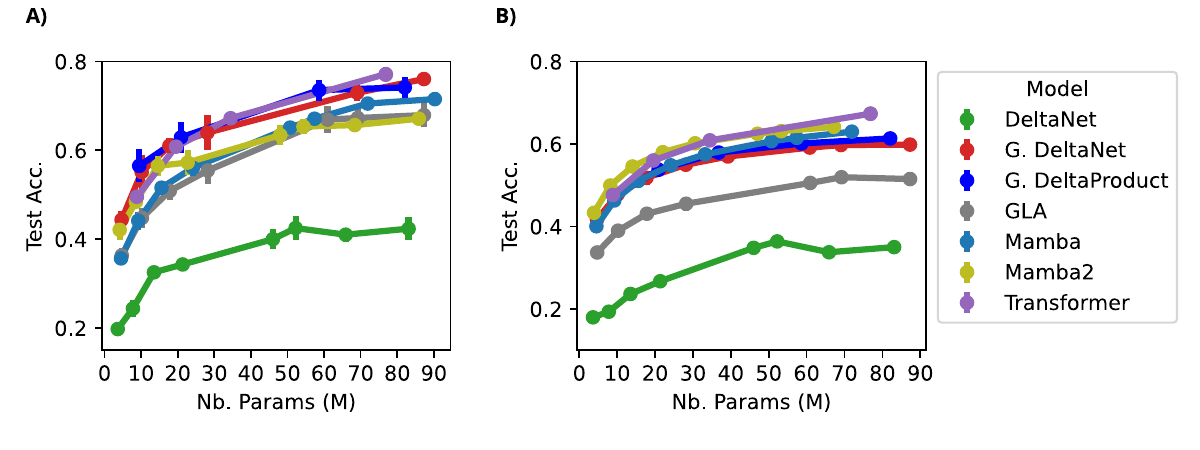}
    \caption{Test accuracy as a function of model size when gaps are introduced between sequence items.
    Results are averaged over 4 runs and reported as mean $\pm$ standard deviation.
    \panel{A} Noisy gaps.
    \panel{B} Non-grammatical gaps.
    \label{fig:acc_selectivity}
    }
\end{figure*}

\section{Experiments}
\label{sec:experiments}

\paragraph{Experimental setup.}

We conduct an extensive experimental evaluation of the following sequence models: Transformer \citep{Vaswani17_attention}, Mamba \citep{Gu23_mamba}, GLA \citep{Yang23_gla}, DeltaNet \citep{Yang24_delta_net}, Gated DeltaNet \citep{Yang24_gated_delta_net}, Mamba2 \citep{Dao24_mamba2}, and Gated DeltaProduct \citep{Siems25_delta_product}, as central models in the exploration of linear recurrent memories (LRMs) encompassing the key features we identified in \cref{sec:models}.
To ensure consistency, we set the expansion factor to $v = 2$ for all applicable models, fix the number of Householder transformations $n_h$ to 4 for the Gated DeltaProduct, and set $d_\text{state}$ to 64 for Mamba and 128 for Mamba2. All other parameters follow the defaults provided in \cite{Yang24_fla}.
The number of attention heads ($N_\text{head}$) is uniformly configured such that $N_\text{head} = d_\text{model}/H_\text{dim}$, where $H_\text{dim}$ is the head size. All architectures consist of 8 layers, with hidden sizes ranging from 256 to 1296 dimensions, resulting in model sizes varying approximately between 10 million and 100 million parameters. Training is performed using the Adam optimizer, with learning rates chosen from the set \{0.0005, 0.001, 0.002\}. Gradient clipping is applied at a threshold of 1.0, and a consistent batch size of 64 is maintained throughout. The learning rate follows a cosine annealing schedule with a warmup period set to either 478 or 1000 steps.

\paragraph{Sequence specifications.} For all experiments except the one shown in \cref{fig:acc_complexity_model_size}A, we use a grammar-based sequence generation task characterized by a vocabulary size \VocabSize of 40, an ambiguity number \NumAmb of 40, an ambiguity depth \AmbDepth of 40, and a transition probability \TransitionProb of $0.001$. This configuration results in a topological entropy (TE) of $2.77$.
Training and testing involve sequences with lengths sampled uniformly between \( T_{\text{min}} = 1 \) and \( T_{\text{max}} = 200 \).
For each experiment, four datasets are generated independently, each containing $200{,}000$ training examples and $1{,}000$ test examples. A separate model is initialized and trained on each dataset. Final accuracy results are averaged over these four datasets to ensure robustness.

\paragraph{Memorization for Disambiguation.}
In the first set of experiments, we assess sequence learning performance in the presence of only a single zero gap between sequence items (\textbf{Task 1}). We examine how performance varies as a function of sequence complexity and model size.
As shown in \cref{fig:acc_complexity_model_size}A, the complexity of the sequence generator, measured by its topological entropy (TE), is varied by sweeping the number of ambiguities \NumAmb over \{20,24,26,30,34,38,40\} and setting the ambiguity depth \AmbDepth = \NumAmb and the vocabulary size \VocabSize = \NumAmb + 1. This systematic increase in ambiguity level and thus sequence complexity leads to a consistent decline in test accuracy across all models.
Models such as Mamba, Mamba2, Gated DeltaNet, Gated DeltaProduct, and Transformer maintain higher accuracy across varying levels of complexity. In contrast, DeltaNet and GLA show the steepest performance decline, attributed to the absence of fast forgetting and complementary gating mechanisms (see \nameref{sec:models}), respectively.
A similar pattern emerges in the model scaling experiment (\cref{fig:acc_complexity_model_size}B). Here, and from now on, we fix the grammar to the complexity TE=2.77 and vary model size by adjusting hidden dimensions. Again, models with complementary gating and fast forgetting mechanisms consistently outperform others across the various parameter scales.
Nonetheless, as in pure memorization benchmarks reported in prior work, this task is not able to distinguish the characteristics of modern LRM models.

\begin{table*}[!h]
\centering
\small
\begin{tabular}{lcccc}
\toprule
\textbf{Model} & \textbf{Nb. Params (M)} & \textbf{Test Acc. (Task 1)} & \textbf{Test Acc. (Task 2)} & \textbf{Test Acc. (Task 3)} \\
\midrule
DeltaNet           & 83  & 0.5 $\pm$ 0.01               & 0.41             $\pm$ 0.01   & 0.35 $\pm$ 0.001\\
GLA                & 87  & 0.75 $\pm$ 0.01              & 0.68             $\pm$ 0.05   & 0.5  $\pm$ 0.01 \\
Mamba              &  90, (71 in task 2)  & \underline{0.89 $\pm$ 0.009} & 0.71             $\pm$ 0.01   & 0.63 $\pm$ 0.08\\
Mamba2             & 85, (67 in task 2)  & \textbf{0.92 $\pm$ 0.007} & 0.67             $\pm$ 0.01   & \underline{0.64 $\pm$ 0.01} \\
Gated DeltaNet     & 87  & 0.87 $\pm$ 0.01 & \underline{0.76  $\pm$ 0.01}  & 0.57 $\pm$ 0.008 \\
Gated DeltaProduct & 82  & 0.87 $\pm$ 0.008 & 0.74            $\pm$ 0.02   & 0.61  $\pm$ 0.01\\
Transformer        & 78  & 0.86 $\pm$  0.005 & \textbf{0.77   $\pm$ 0.01}  & \textbf{0.67 $\pm$ 0.07} \\
\bottomrule
\end{tabular}
\caption{Test accuracy at the largest parameter size for each model for the various tasks.
The best model is in bold while the second best is underlined.
Results are averaged over 4 runs and reported as mean $\pm$ standard deviation.
}
\label{tab:scaling_law_results}
\end{table*}

\paragraph{Selectivity assessment.}
To address this evaluation gap, we analyze the influence of noise gaps (\textbf{Task 2}) and non-grammatical gaps (\textbf{Task 3}) on sequence learning performance. The results are reported in \cref{fig:acc_selectivity}.
The introduction of noise gaps between sequence items reduces overall test accuracy compared to the \textbf{Task 1} setup (during training, the number of gaps is controlled by setting $n_{\text{min}}=0$ and $n_{\text{max}}=10$, and fixed to 10 at test time, with noise amplitude in the range $[0, \gamma]$, where $\gamma$ is set to 0.2).
This highlights the difficulty these models face in discarding the noise token (\cref{fig:acc_selectivity}A).
The Transformer, followed by Gated DeltaNet and Gated DeltaProduct, achieves the highest accuracy across model sizes. DeltaNet displays poor performance. This can be attributed to the absence of a fast forgetting mechanism, which results in memory overload from noisy gaps. GLA, Mamba, and Mamba2 achieve intermediate performance compared to the other models, indicating that channel mixing in the state update plays an important role when discarding easily identifiable gaps.
\par
In the context-aware selectivity experiment of \textbf{Task 3}, we insert non-grammatical gap tokens of fixed duration 1 after each sequence token with probability $p_\text{G}$. During training, we set $p_\text{G}=0.1$, which allows the model to learn the grammar in a less challenging setting. At test time, we use $p_\text{G}=1$ (\cref{fig:acc_selectivity}B).
The first observation is that non-grammatical gap insertion leads to a general decrease in performance and a more differentiated ranking of models. DeltaNet continues to perform poorly due to its weak selective gating, while the performance of GLA drops relative to \textbf{Task 2}.
Mamba2, followed by Mamba, achieves the strongest performance across scales, pointing to the importance of the complementary gating mechanism (see \nameref{sec:models}) for handling more complex distractors.
In contrast, Gated DeltaProduct and Gated DeltaNet rank lower than in \textbf{Task 2}, suggesting that in-state channel mixing is less critical in this setting.
The diminished role of channel mixing is also consistent with results on natural language processing (NLP) tasks. For example, in Gated DeltaNet \cite{Yang24_gated_delta_net}, a considerable improvement over DeltaNet is observed on language modeling, while Mamba2 also achieves very similar performance to Gated DeltaNet despite lacking channel-mixing dynamics in its state update.
This could potentially be due to the MLP layers already implementing the channel-mixing operations between recurrent blocks, thereby making additional channel mixing in the recurrent update largely redundant.
We further attribute the strong performance of Mamba and Mamba2 to their exponential gating and logarithmic reparameterization, which have been shown to improve training stability in recurrent models \citep{zucchet2024recurrent}. Mamba2, in particular, benefits from a larger state dimension at a comparable parameter budget, enabled by its low-parameter state transition \citep{Dao24_mamba2}. Overall, these results support the view that the proposed context-aware selectivity task serves as a practical small-scale proxy for benchmarking LRMs in large language modeling.


\paragraph{Gap generalization.}

\begin{figure}[!h]
    \centering
    \includegraphics[width=\linewidth]{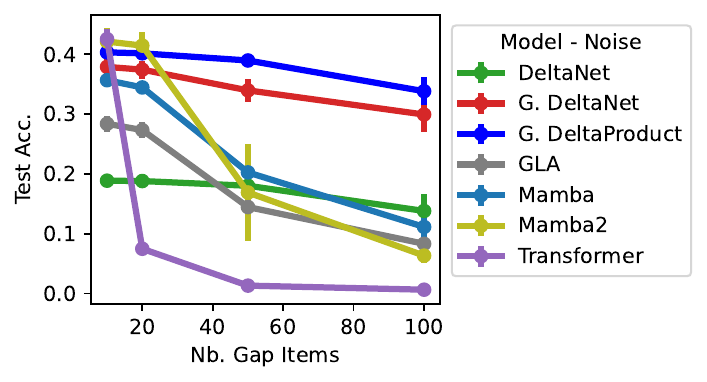}
    \caption{Generalization accuracy as a function of the number of gap items.
     Results are averaged over 4 runs and reported as mean $\pm$ standard deviation.}
    \label{fig:gap_generalization}
\end{figure}

Finally, following \textbf{Task 4}, we evaluate how well models generalize to larger temporal gaps during inference. 
The models are trained using the same training methodology as \textbf{Task 2}, but are tested on sequences with an increased number of noise gap duration between sequence tokens, going up to 100 items.
\cref{fig:gap_generalization} shows that as the number of gap items increases, test accuracy drops across all models.
This may reflect increasing difficulty in retaining information over longer temporal gaps, or corruption of the model's internal dynamics.
Gated DeltaNet and Gated DeltaProduct retain higher accuracy under increasing gap sizes, suggesting that channel-mixing of the delta rule plays a crucial role for generalization, at least on noisy gaps. In contrast, Transformer performance degrades rapidly, approaching chance levels, confirming its limited extrapolation ability known from previous studies \cite{Deletang22_neural}.
We note, however, that techniques such as curriculum learning \cite{kim2024strategic} have already been proposed to improve this capability in LLMs.
DeltaNet and Mamba models show moderate resilience, while Mamba2 exhibits a steeper drop, consistent with its lighter parametrization design. These results emphasize the different requirements in model features for length generalization (\textbf{Task 4}) with respect to memorization (\textbf{Task 1}) and selectivity (\textbf{Tasks 2 and 3}).

\paragraph{Computational Cost.}

Finally, we observe that the additional computational cost of Gated DeltaProduct is not justified for the tasks considered here, as it never shows a significant advantage over cheaper alternatives such as Gated DeltaNet or Mamba2. By contrast, our results support the use of lighter models such as Mamba2, whose smaller state-update parameterization leads to better scaling behavior, as evidenced in \textbf{Task 1} and \textbf{Task 3}.

\section{Discussion and conclusion}

We presented the SelectivBench, specifically designed to assess selectivity in sequence learning models. By combining rule-based sequence generation with controlled complexity and structured disruptions, this newly introduced set of benchmark tasks enables systematic evaluation of models' selective ability to memorize relevant input, while ignoring gaps that violate transition rules. Unlike prior synthetic tasks, our framework allows fine-grained control over sequence complexity and assesses the relevance of various gating features of recent linear recurrent models.
Our experiments shed light on key architectural components: gating and rapid forgetting support effective recall; in-state channel mixing, though not required for selectivity, is crucial for generalization; and softmax attention remains dominant due to its expanding memory capacity.
A direction for future work is to test the proposed framework on sequence generators with higher complexity and from other grammar classes, including context-free and context-sensitive languages \cite{Deletang22_neural}.

\section{Acknowledgements}
This work was funded by the Federal Ministry of Research, Technology, and Space, Germany (project NEUROTEC-II grant no. 16ME0398K and 16ME0399).


\end{document}